\documentclass{article}

\usepackage{arxiv}
\usepackage{amssymb}
\usepackage{amsmath}
\usepackage{bm}
\usepackage{subfig}
\usepackage{wrapfig}
\usepackage{lipsum}     
\usepackage{float} 
\usepackage{graphicx}           
\usepackage{url} 
\usepackage{soul}
\usepackage{enumitem}
\usepackage{lineno}
\usepackage{diagbox}
\usepackage{color}
\usepackage{booktabs} 

\usepackage[ruled,vlined,linesnumbered,noresetcount]{algorithm2e}

\title{Output-Weighted Sampling for Multi-Armed Bandits with Extreme Payoffs}

\author{
  Yibo Yang \\
  Department of Mechanical Engineering \\
  and Applied Mechanics\\
  University of Pennsylvania\\
  Philadelphia, PA 19104 \\
  \texttt{ybyang@seas.upenn.edu} \\
   \And
    Antoine Blanchard \\
  Department of Mechanical Engineering \\
  Massachusetts Institute of Technology\\
  Cambridge, MA 02139 \\
  \texttt{ablancha@mit.edu} \\
   \And
    Themistoklis Sapsis \\
  Department of Mechanical Engineering \\
  Massachusetts Institute of Technology\\
  Cambridge, MA 02139 \\
  \texttt{sapsis@mit.edu} \\
   \And
  Paris Perdikaris \\
  Department of Mechanical Engineering \\
  and Applied Mechanics\\
  University of Pennsylvania\\
  Philadelphia, PA 19104 \\
  \texttt{pgp@seas.upenn.edu} \\
}

\begin{document}
\maketitle

\begin{abstract}
We present a new type of acquisition functions for online decision making in multi-armed and contextual bandit problems with extreme payoffs. Specifically, we model the payoff function as a Gaussian process and formulate a novel type of upper confidence bound (UCB) acquisition function that guides exploration towards the bandits that are deemed most relevant according to the variability of the observed rewards. This is achieved by computing a tractable likelihood ratio that quantifies the importance of the output relative to the inputs and essentially acts as an \textit{attention mechanism} that promotes exploration of extreme rewards. We demonstrate the benefits of the proposed methodology across several synthetic benchmarks, as well as \textcolor{black}{two} realistic examples involving noisy sensor network data \textcolor{black}{(specifically, temperature and air quality measurements)}. Finally, we provide a JAX library for efficient bandit optimization using Gaussian processes. 
\end{abstract}


\section{Introduction}\label{sec:intro}
Online decision making defines an important branch of modern machine learning in which uncertainty quantification plays a prominent role. In most stochastic optimization settings, evaluating the unknown function is expensive, hence new information needs to be acquired judiciously.

Classical applications include recommendation systems for articles and products, where the goal is to maximize the total revenue of the product maker given limited user feedback \cite{li2010contextual,kawale2015efficient}; control and reinforcement learning, where the reward is obtained after a sequence of experiments or actions and the objective is not only to obtain optimal rewards but also avoid the potentially negative effects of uncertainty \cite{dearden1998bayesian,osband2016deep,azizzadenesheli2018efficient,li2019adversarial}; environment monitoring, where sensor data is used to identify areas of interest as in traffic flow estimation \cite{srinivas2009gaussian} and room temperature monitoring \cite{krause2011contextual}; and optimal design of expensive experiments \cite{sacks1989design,saha2008optimum}.  
 
More recently, new applications have appeared beyond machine learning, including optimal sampling in cardiac electrophysiology and bio-engineering \cite{costabal2019multi,sahli2020physics}, multi-fidelity design of experiments \cite{forrester2007multi,sarkar2019multifidelity}, hyper-parameter tuning in high-dimensional design spaces \cite{shan2010survey,perdikaris2016multifidelity, bouhlel2016improving}, and prediction of extreme events in complex dynamical systems \cite{wan2018data, mohamad2018sequential}.

Many of these applications can be formulated as multi-armed bandit problems, for which effective sampling algorithms exist  \cite{auer2002using,srinivas2009gaussian,chu2011contextual,krause2011contextual,schaul2015prioritized,osband2016deep}. These algorithms are generally characterized by two key ingredients. First, they involve building a model for the latent payoff function given scarce and possibly noisy observations of past rewards. To enable effective sampling and exploration of the decision space, uncertainty in the model predictions needs to be accounted for in the predictive posterior distribution of the latent payoffs, which can be obtained via either a frequentist or a Bayesian approach. The second critical ingredient pertains to designing a data acquisition policy that can leverage the model predictive uncertainty to effectively balance the trade-off between exploration and exploitation while ensuring a consistent asymptotic behavior for the cumulative regret.

\subsection{Previous Work}

Multi-armed bandit problems provide a general setting for developing online decision-making algorithms and rigorously studying their performance. Early research in this setting includes the celebrated $\epsilon$-greedy algorithm \cite{schaul2015prioritized}, where random exploration is introduced with a small probability $\epsilon$ to prevent the algorithm from focusing on local sub-optimal solutions. Despite its widespread applicability, $\epsilon$-greedy algorithms employ a heuristic treatment of uncertainty, and often require careful tuning in order to prevent sub-optimal exploration. 

To this end, the upper confidence bound (UCB) policy \cite{agrawal1995sample,auer2002using} was proposed to provide a natural way to estimate sub-optimal choices using a model's predictive posterior uncertainty. However, the original UCB formulation does not take into account correlations between different bandits in a multi-armed setting and, therefore, typically requires a large number of datapoints to be collected before convergence can be observed. Variants of the UCB algorithm have been adapted to the contextual bandit setting with linear payoffs, where the payoff function is modeled via Bayesian linear regression \cite{chu2011contextual}. Gaussian process models have also been employed to account for correlated payoffs, and the corresponding GP-UCB criteria have shown great promise in data-scarce and ``cold start'' scenarios \cite{dani2008stochastic,srinivas2009gaussian, krause2011contextual}.  

Thompson sampling \cite{thompson1933likelihood,russo2018tutorial} provides an alternative approach to balancing the exploration--exploitation trade-off that only requires access to posterior samples of a parametrized payoff function. Although the algorithm was largely ignored at the time of its inception by Thompson \cite{thompson1933likelihood}, the results of Chapelle \& Li \cite{chapelle2011empirical} have initiated a wave of resurgence, leading to significant advances in applications (e.g., recommendation systems \cite{kawale2015efficient}, hyper-parameter optimization \cite{kandasamy2018parallelised}, reinforcement learning \cite{dearden1998bayesian,osband2016deep,azizzadenesheli2018efficient}), as well as theoretical analyses (e.g., optimal regret bounds \cite{kaufmann2012thompson,leike2016thompson,russo2016information}).  More recently, Bayesian deep learning models have been considered  \cite{graves2011practical} for modeling more complex and high-dimensional payoff functions.  However, their effectiveness, interpretability, and convergence behavior are still under investigation \cite{riquelme2018deep}.

\subsection{Our contributions}

\paragraph{Primary contribution.} All aforementioned approaches have enjoyed success across various applications, however they lack a mechanism for distinguishing and promoting the input/context variables that have the greatest influence on the observed payoffs. Short of such mechanism, regions in the decision space that may have negligible effect on the payoffs will still be sampled as long as they are uncertain. As we will demonstrate, this undesirable behavior can have a deteriorating impact on convergence, and this effect is exacerbated in the presence of extreme payoffs (i.e., situations in which a small number of bandits yield rewards  significantly greater than the rest of the bandit population). 

Motivated by the recent findings in  \cite{sapsis2020output,blanchard2020output,blanchard2020bayesian}, we introduce a novel UCB-type objective for online decision making in multi-armed and contextual bandit problems that can overcome the aforementioned pathologies. This is achieved by introducing an importance weight to effectively promote the exploration of ``heavy-tailed'' (i.e., rare and extreme) payoffs. We show how such importance weight can be derived from a likelihood ratio that quantifies the relative importance between inputs/contexts and observed rewards, introducing an effective \textit{attention mechanism} that favors exploration of bandits with unusually large rewards over bandits associated with frequent, average payoffs. This output-weighted approach has been shown to outperform classical acquisition functions in active learning \cite{blanchard2020informative} and Bayesian optimization \cite{blanchard2020bayesian} tasks, and here we set sail for the first time into investigating its effectiveness in online decision making tasks, with a specific focus on multi-armed and contextual bandit problems subject to extreme payoffs.

\paragraph{Comparison to previous work.} We demonstrate the effectiveness of the proposed methodology across a collection of synthetic benchmarks, as well as \textcolor{black}{two} realistic examples involving noisy sensor network data \textcolor{black}{(specifically, temperature and air quality measurements)}. In all cases, we provide comprehensive quantitative comparisons between the proposed output-weighted sampling criterion and the most widely-used criteria in current practice, including the UCB \cite{auer2002using}, GP-UCB \cite{srinivas2009gaussian}, Thompson sampling \cite{thompson1933likelihood, chapelle2011empirical}, and expected improvement \cite{vazquez2010convergence} methods. 

\paragraph{Secondary contributions.} We have developed an open-source Python package for bandit optimization using Gaussian processes\footnote{\url{https://github.com/PredictiveIntelligenceLab/jax-bandits}}. Our implementation leverages the high-performance package JAX \cite{jax2018github} and thus enables (a) gradient-based optimization of the proposed output-weighted sampling criteria for general Gaussian process priors, (b) the use of GPU acceleration, and (c) scalability and parallelization across multiple computing nodes. This package can be readily used to reproduce all data and results presented in this paper.

\section{Methods}\label{sec:Methods}

\subsection{Multi-armed bandits}\label{sec:IntroMultiArmedBandits}

The multi-armed bandit problem is a prototypical paradigm for sequential decision making.  The decision set consists of a discrete collection of $M$ arms where the $i$th arm may be associated with some contextual information $\mathbf{x}_i\in \mathbb{R}^d$.  Pulling arm $i$ produces a reward $y \in \mathbb{R}$ which is determined by some unknown latent function 
\begin{equation}
y_t = f(\mathbf{x}_i) + \epsilon_t,
\end{equation}
where $\epsilon_t \sim \mathcal{N}(0,\,\sigma_n^{2})$ accounts for observation noise.  

At each round $t$, we select an arm $i$ and obtain a reward $y_t$. The goal of sequential decision making is to find a strategy for bandit selection that maximizes the total reward $\sum_{t = 1}^{T}y_t$ for a given budget $T$. In other words, the goal is to first identify the bandits that provide the best rewards,
\begin{equation}\label{equ:BestArms}
    \mathbf{x}^{*} = \arg\max_{\mathbf{x}} f(\mathbf{x}),
\end{equation}
using as few arm pulls as possible, and then to keep on exploiting these optimal bandits to maximize the total reward.

As an alternative metric of success, it is useful to consider the simple regret $r_t = f(\mathbf{x}^{*}) - f(\mathbf{x})$, as maximizing the total reward is essentially equivalent to minimizing the \textit{cumulative} regret 
\begin{equation}\label{equ:TotalRegrets}
    R_T = \sum_{t = 1}^{T}r_t.
\end{equation}
The holy grail of online decision making is to design an effective no-regret policy satisfying
\begin{equation}
\lim_{T\rightarrow\infty}\frac{R_T}{T} = 0.
\end{equation}

\subsection{Gaussian Processes}\label{sec:GP}

Gaussian process (GP) regression provides a flexible probabilistic framework for modeling nonlinear black-box functions  \cite{rasmussen2003gaussian}. Given a dataset $\mathcal{D} = \{(\mathbf{x}_{i}, y_{i})\}_{i=1}^N$ of input--output pairs (i.e, context--reward pairs), and an observation model of the form $y = f(\mathbf{x})+ \epsilon$, the goal is to infer the latent function $f$ as well as the unknown noise variance $\sigma_n^{2}$  corrupting the observations.

In GP regression, no assumption is made on the form of the latent function $f$ to be learned; rather, a prior probability measure is assigned to every function in the function space. Starting from a zero-mean Gaussian prior assumption on $f$, 
\begin{equation}
f(\mathbf{x}) \sim \text{GP}(\mathbf{0},
k(\mathbf{x},\mathbf{x'};\bm{\theta})),
\end{equation}
the goal is to identify an optimal set of hyper-parameters  $\bm{\Theta}=\{\bm{\theta}, \sigma_{n}^{2}\}$, and then use the optimized model to predict the rewards of unseen bandits.  The covariance function $k(\mathbf{x},\mathbf{x'};\bm{\theta})$ plays a key role in this procedure as it encodes prior belief or domain expertise one may have about the underlying function $f$. In the absence of any domain-specific knowledge, it is common to assume that $f$ is a smooth continuous function and employ the squared exponential covariance kernel with automatic relevance determination (ARD) which  accounts for anisotropy with respect to each input variable \cite{rasmussen2003gaussian}.

Unlike previous works \cite{srinivas2009gaussian, krause2011contextual}, we do not assume that the payoff function $f$ actually comes from a GP prior or that it has low RKHS norm. Instead, we compute an optimal set of hyper-parameters at each round $t$ by minimizing the negative log-marginal likelihood of the GP model \cite{rasmussen2003gaussian}. In our setup, the likelihood is Gaussian and can be computed analytically as
\begin{align} \label{equ:NLML}
\mathcal{L}(\bm\Theta) &=  \frac{1}{2}\log|\mathbf{K}+\sigma_{n}^{2}\mathbf{I}| \nonumber\\
    &+ \frac{1}{2}\mathbf{y}^\mathsf{T} (\mathbf{K}+\sigma_{n}^{2}\mathbf{I})^{-1}\mathbf{y} + \frac{N}{2}\log (2\pi),
\end{align}
where $\mathbf{K}$ is an $N\times N$ covariance matrix constructed by evaluating the kernel function on the input training data $\mathbf{X}$. The minimization problem is solved with an L-BFGS optimizer with random restarts \cite{liu1989limited}.

Once the GP model has been trained, the predictive distribution at any given bandit $\mathbf{x}$ can be computed by conditioning on the observed data:
\begin{equation}
    p(\mathbf{y}\mid\mathbf{x},\mathcal{D})\sim\mathcal{N}(\mu(\mathbf{x}), \sigma^2(\mathbf{x})),
\end{equation}
where
\begin{subequations}
\begin{gather}
    \mu(\mathbf{x})  =  k(\mathbf{x}, \mathbf{X}) (\mathbf{K} + \sigma_{n}^{2}\mathbf{I})^{-1}\mathbf{y}, \label{equ:posterior_mean} \\
    \sigma^2(\mathbf{x}) =  k(\mathbf{x}, \mathbf{x}) - k(\mathbf{x}, \mathbf{X}) (\mathbf{K} + \sigma_{n}^{2}\mathbf{I})^{-1} k(\mathbf{X},\mathbf{x}). \label{equ:posterior_variance}
\end{gather}
\end{subequations}
Here, $\mu(\mathbf{x})$ can be used to make predictions and $\sigma^2(\mathbf{x})$ to quantify the associated uncertainty.

\subsection{Online decision making}\label{sec:AcquisitionFunction}

A critical ingredient in online decision making is the choice of the \textit{acquisition function}, which effectively determines  which bandits the algorithm should try out and which ones to ignore \cite{srinivas2009gaussian,krause2011contextual}. A popular choice of acquisition function is the ``vanilla'' upper confidence bound (V-UCB),
\begin{equation} \label{eq:vucb}
    a_\text{V-UCB}(\mathbf{x}) = \mu(\mathbf{x}) + \kappa\sigma(\mathbf{x}),
\end{equation}
and the closely-related GP-UCB criterion \cite{srinivas2009gaussian},
\begin{equation} \label{eq:gpucb}
    a_\text{GP-UCB}(\mathbf{x}) = \mu(\mathbf{x}) + \beta_t^{1/2}\sigma(\mathbf{x}),
\end{equation}
where $\kappa$ and $\beta_t = 2\log(|D|t^2\pi^2/(6\delta))$ are parameters that aim to balance exploration and exploitation. ($|D|$ is the number of bandits in the absence of context, and the dimension of the context otherwise.) In V-UCB, $\kappa$ is typically considered constant, while in GP-UCB, $\beta_t$ depends on the round $t$ and comes with convergence guarantees when the payoff function is not too complex \cite{srinivas2009gaussian}. 

In this work we also consider the expected improvement,
\begin{equation} \label{eq:ei}
    a_\text{EI}(\mathbf{x}) = \sigma(\mathbf{x})[\lambda(\mathbf{x})\Phi(\lambda(\mathbf{x}))+\phi(\lambda(\mathbf{x}))],
\end{equation}
whose convergence properties have been well studied \cite{vazquez2010convergence}, as well as Thomson sampling,
\begin{equation} \label{eq:ts}
    a_\text{TS}(\mathbf{x}) = \Tilde{y}(\mathbf{x}),
\end{equation}
also known to deliver competitive results in practice
\cite{chapelle2011empirical,agrawal2012analysis,riquelme2018deep}. In \eqref{eq:ei}, we have defined $\lambda(\mathbf{x}) = (\mu(\mathbf{x}) - y^* -  \xi)/\sigma(\mathbf{x})$, with $y^*$ the best reward recorded so far and $\xi$ a user-specified parameter controlling the exploration--exploitation trade-off.  The quantity $\tilde{y}(\mathbf{x})$ in \eqref{eq:ts} denotes a random sample drawn from the posterior distribution of the GP model, that is, $\tilde{y}(\mathbf{x})\sim\mathcal{N}(\mu(\mathbf{x}), \sigma^2(\mathbf{x}))$.

The goal in bandit optimization is to determine the best bandit to try next by maximizing the acquisition function:
\begin{equation}\label{equ:AcquisitionPoint}
    \mathbf{x}_{t+1} = \operatorname*{arg\,max}_{\mathbf{x}} a(\mathbf{x};\mathcal{D}),
\end{equation}
where $a$ can be any of \eqref{eq:vucb}, \eqref{eq:gpucb}, \eqref{eq:ei}, or \eqref{eq:ts}, and $\mathcal{D}$ contains all the observed context--reward pairs up to round $t$.

\subsection{Output-weighted sampling}\label{sec:LWAcquisitionFunction}
\begin{figure*}[t]
\centering
\includegraphics[width=0.95\textwidth,clip=true,trim=10 16 0 16]{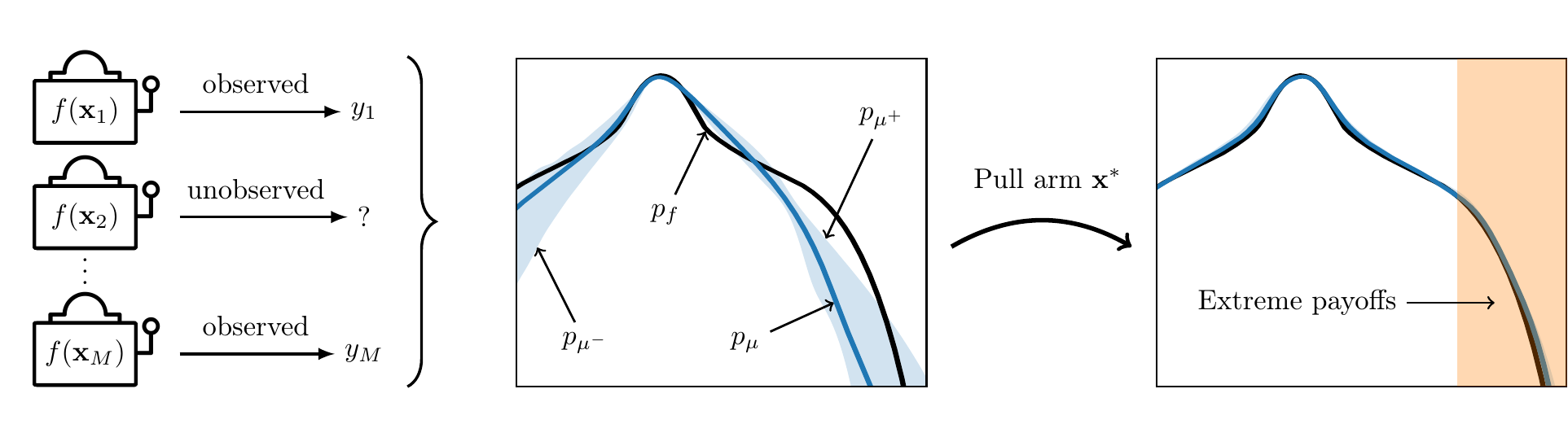}
\caption{Sketch of the acquisition scheme from which the likelihood ratio is derived.  The best next bandit $\mathbf{x}^*$ maximizes the reduction of the uncertainty in the tails of the payoff distribution (quantified by the log-difference between $p_{\mu^+}$ and $p_{\mu^-}$).}
\label{fig:lik}
\end{figure*}

Blanchard \& Sapsis \cite{blanchard2020bayesian} recently introduced an efficient and minimally intrusive approach for accelerating the stochastic optimization process in cases where certain regions of the input space have a considerably larger impact on the output of the latent function than others (i.e., extreme payoffs in the bandit problem) by incorporating a sampling weight into several of the acquisition functions commonly used in practice.  The sampling weight, referred to as the ``likelihood ratio'', was derived from a heavy-tail argument whereby the best next input point to visit is selected so as to most reduce the uncertainty in the \textit{tails} of the output statistics where the extreme payoffs ``live'' (Figure \ref{fig:lik}).

The likelihood ratio is defined as
\begin{equation}\label{equ:Likelihood_ratio}
    w(\mathbf{x}) = \frac{p_{\mathbf{x}}(\mathbf{x})}{p_{\mu}(\mu(\mathbf{x}))}
\end{equation}
and was derived in \cite{blanchard2020bayesian}.  Here, $p_{\mathbf{x}}(\mathbf{x})$ is a prior distribution that can be used to distill prior beliefs about the importance of each bandit or environmental conditions.  In this work we assume that no such prior information is available and treat every bandit equally by specifying a uniform prior, $p_{\mathbf{x}}(\mathbf{x}) = 1$ for all $\mathbf{x}$.  The term $p_{\mu}(\mu(\mathbf{x}))$ denotes the output density of the payoff function and plays an important role to determine the best arms to pull. 

The intuition behind the likelihood ratio is as follows.  Assuming enough data has been collected, the GP posterior mean $\mu(\mathbf{x})$ provides a good estimation about the distribution of rewards for the bandits.  Bandits with unusually large rewards are associated with small values of $p_{\mu}$, while bandits with frequent, average rewards are associated with large values of $p_{\mu}$.  Because the output density $p_{\mu}$ appears in the denominator of \eqref{equ:Likelihood_ratio}, the likelihood ratio assigns more weight to bandits with extreme payoffs.  As such, the likelihood serves as an \textit{attention mechanism} which encourages the algorithm to explore bandits whose rewards are thought to be abnormally large, while penalizing the other mediocre bandits by assigning them small weights. 

To obtain a well-behaved (i.e., smooth and bounded) analytical approximation of the likelihood ratio, we use a Gaussian mixture model,
\begin{equation}\label{equ:GMMApproximation}
    w(\mathbf{x}) \approx \sum_{k=1}^{n_\textit{GMM}} \alpha_k \mathcal{N}(\mathbf{x}; \mathbf{\gamma}_k, \bm{\Sigma}_k),
\end{equation}
where $\mathcal{N}(\mathbf{x}; \mathbf{\gamma}_k, \bm{\Sigma}_k)$ denotes the $k$th component of the mixture model with mean $\mathbf{\gamma}_k$ and covariance $\bm{\Sigma}_k$. The resulting output-weighted acquisition function for the bandit optimization problem is given by
\begin{equation}\label{equ:LWAcquisition}
    a_\text{LW-UCB}(\mathbf{x}) = \mu(\mathbf{x}) + \kappa w(\mathbf{x})\sigma(\mathbf{x}),
\end{equation}
where the subscript ``LW-UCB'' stands for ``likelihood-weighted UCB''.  Equation \eqref{equ:LWAcquisition} is subject to the same bandit-selection policy as the acquisition functions in Section \ref{sec:AcquisitionFunction}:
\begin{equation}\label{equ:LWAcquisitionPoint}
    \mathbf{x}_{t+1} = \arg\max_{\mathbf{x}} a_\text{LW-UCB}(\mathbf{x};\mathcal{D}).
\end{equation}
In general, the minimization problem can be efficiently solved with an L-BFGS optimizer with random restarts \cite{liu1989limited}, where the gradient of the acquisition function with respect to the inputs $\mathbf{x}$ can be computed analytically for the squared exponential covariance kernel \cite{blanchard2020bayesian}, or using automatic differentiation \cite{baydin2015automatic} for more general kernel choices. The workflow for output-weighted sampling with LW-UCB is summarized in Algorithm \ref{algorithm:LW_UCB}. 
\begin{algorithm}
\SetAlgoLined
\textbf{Input:} Small initial dataset $\mathcal{D} = \{(\mathbf{x}_{i}, y_{i})\}_{i=1}^n$\;
\While{$t < T$}{
    Fit GP model to dataset $\mathcal{D}$ using \eqref{equ:NLML} and obtain posterior mean \eqref{equ:posterior_mean} and variance \eqref{equ:posterior_variance}\;
    Compute likelihood ratio \eqref{equ:Likelihood_ratio} and fit  Gaussian mixture model \eqref{equ:GMMApproximation} to it \;
    Select best next bandit $\mathbf{x}_{t+1}$ by maximizing \eqref{equ:LWAcquisition}\;
    Collect new reward $y_{t+1} = f(\mathbf{x}_{t+1} ) + \epsilon_{t+1}$ and append $(\mathbf{x}_{t+1}, y_{t+1})$ to dataset $\mathcal{D}$\;
 }
 \caption{The LW-UCB algorithm.}
 \label{algorithm:LW_UCB}
\end{algorithm}

\section{Results}
\label{sec:Results}

In all numerical studies considered in this work, we initialize the algorithm with $n=3$ random input--output pairs and compare the performance of EI, TS, V-UCB, GP-UCB, and LW-UCB.  Our metric of success is the log-cumulative regret over time. Unless otherwise indicated, we conduct a series of $100$ random experiments, each with a different choice of initial data, and report the median of the metric of interest.  Variability across experiments is quantified using the median absolute deviation.

\subsection{Synthetic benchmarks}\label{sec:classic_functions}

We demonstrate the performance of LW-UCB for three synthetic test functions.  We consider 2500 bandits arranged on a uniform $50\times 50$ grid with rewards being given by the value of the test function at that point in the domain.  The rewards collected during optimization are corrupted by small Gaussian noise with $\sigma_n = 10^{-4}$. 

\begin{figure*}[t]\label{fig:1}
\centering
\subfloat[\label{fig:1a}Cosine function]{\includegraphics[width=\textwidth]{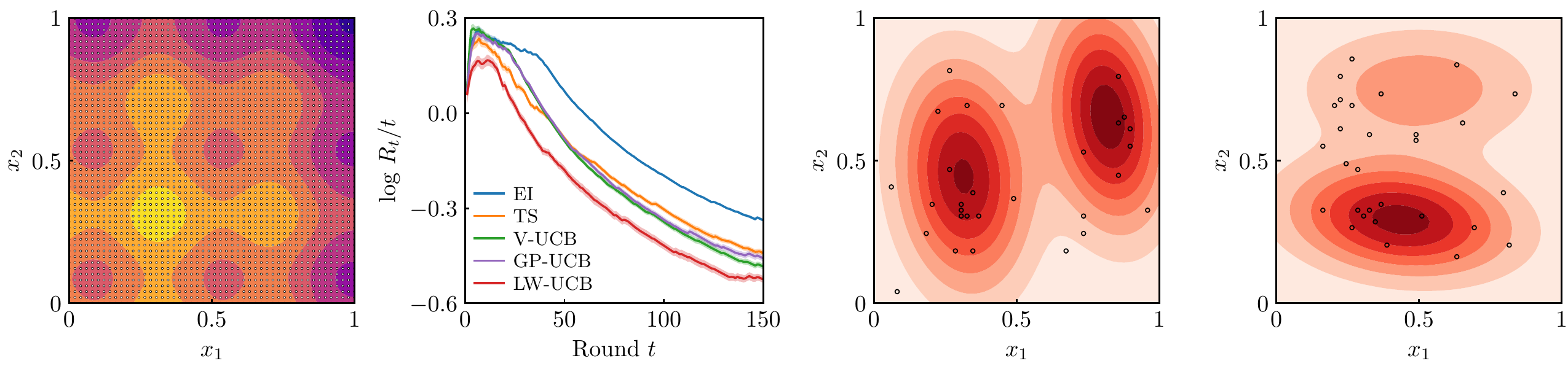}}

\subfloat[\label{fig:1b}Michalewicz function]{\includegraphics[width=\textwidth]{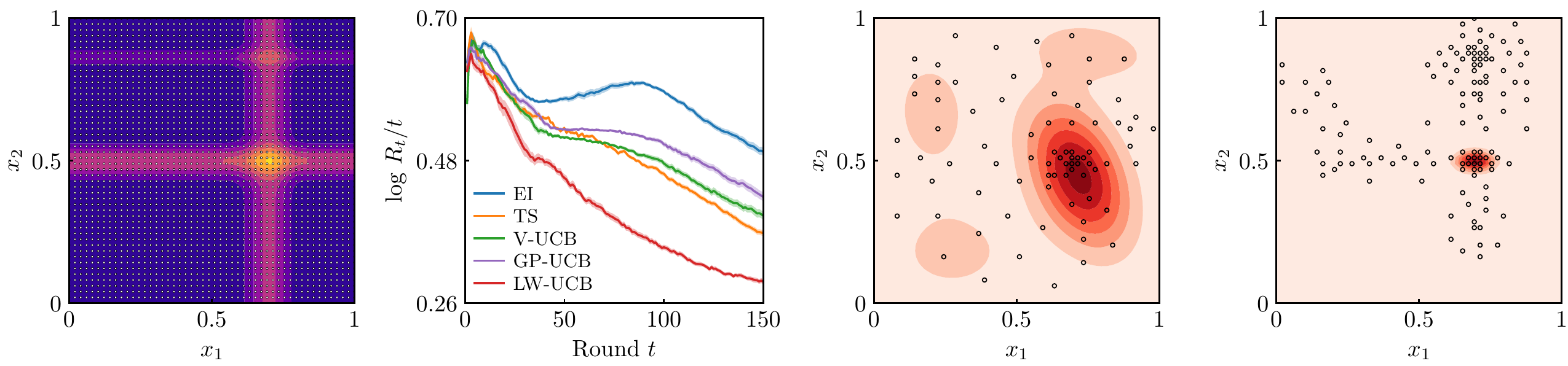}}

\subfloat[\label{fig:1c}Modified Michalewicz function]{\includegraphics[width=\textwidth]{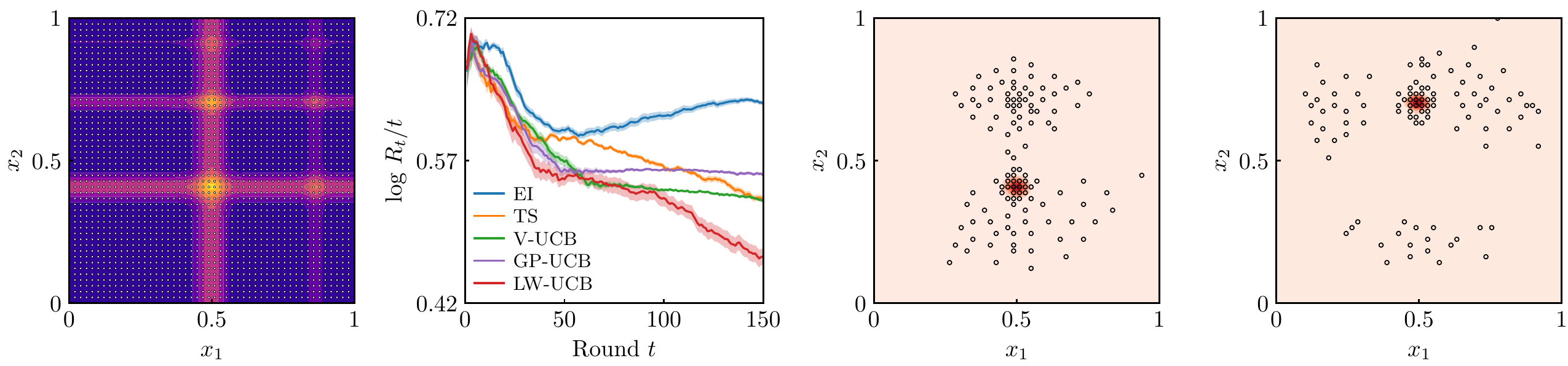}}

\caption{{\em Synthetic benchmarks.} From left to right: locations of the bandits (white circles) and associated rewards (background color); cumulative regret for various acquisition functions; for two representative trials of LW-UCB, distribution of the likelihood ratio (background color) learned by the GP model from the visited bandits (open circles) after $t=150$ rounds.}
\label{fig:Cosine}
\end{figure*}

We begin with the Cosine function from \cite{azimi2010batch},
\begin{equation}\label{equ:Cosine_function}
	f(\mathbf{x}) = 1 - [u^2 + v^2 - 0.3\cos(3\pi u) - 0.3\cos(3\pi v)]
\end{equation}
where $u = 1.6x_1 - 0.5$, $v = 1.6x_2 - 0.5$, and $\mathbf{x} \in [0,1]^2$. For $n_\textit{GMM} = 2$, Figure \ref{fig:1a} shows that LW-UCB performs better than the other methods as it leads to faster identification of the best bandit.  Moreover, Figure \ref{fig:1a} demonstrates how the likelihood ratio highlights the importance of the bandits and favors exploration of those with the highest rewards. We also note the subpar performance of EI, consistent with the discussion in \cite{NIPS2017_b19aa25f}.

Next, we consider the Michalewicz function \cite{azimi2010batch},
\begin{equation}\label{equ:Michalewicz_function}
f(\mathbf{x}) = \sin(\pi x_1)\sin^{20}(\pi x_1^2) + \sin(\pi x_2)\sin^{20}(2 \pi x_2^2),
\end{equation}
with $\mathbf{x} \in [0,1]^2$.  This function is more challenging than the Cosine function as it exhibits large areas of ``flatland'' (i.e., many mediocre bandits) and a very deep and narrow well located slightly off center (i.e., rare bandits with extreme payoffs).  For $n_\textit{GMM} = 4$, Figure \ref{fig:1b} shows that LW-UCB outperforms the competition by a substantial margin.  Figure \ref{fig:1b} also makes it visually clear that the likelihood ratio assigns more weight to the best bandits.  Interestingly, we have found that the likelihood ratio sometimes discovers a broader area where other sub-optimal solutions are also captured.

For an even more challenging test case, we introduce a modified version of the Michalewicz function which features multiple small ``islands'' associated with extreme payoffs.  Specifically, the function
\begin{equation}\label{equ:Modified_Michalewicz_function}
	f(\mathbf{x}) = \sin(\pi x_1)\sin^{20}(2 \pi x_1^2) + \sin(\pi x_2)\sin^{20}(3 \pi x_2^2)
\end{equation}
has six extreme local minima and a number of steep valleys in the domain $\mathbf{x}\in[0,1]^2$, making it quite difficult for the algorithms to identify the best bandits.  Figure \ref{fig:1c} shows that despite the added difficulty, LW-UCB again exhibits outstanding convergence behavior, with the other acquisition functions struggling to identify the best bandits and therefore yielding poor performance. We also note that the likelihood ratio not only emphasizes the best area for rewards but is also able to identify sub-optimal solutions of somewhat lesser importance, demonstrating the ability of our approach to provide a good balance between exploration and exploitation. 

To investigate the effect of the likelihood ratio on runtime, we record the time required to perform one iteration of the Bayesian algorithm.  (This includes training the GP model, computing the likelihood ratio and the GMM approximation for LW-UCB, and optimizing the acquisition function.)  Consistent with \cite{blanchard2020bayesian}, Table \ref{tab:time_cost} shows that the runtimes for LW-UCB are on the same order of magnitude as the other criteria. The additional cost is attributable to the computation and sampling of the likelihood ratio, and presumably can be alleviated using recent advances in sampling methods for GP posteriors \cite{wilson2020efficiently}.

\begin{table*}[!ht]
    \centering
    \caption{Single-iteration runtime (in seconds) averaged over ten experiments.}\label{tab:time_cost}
    \begin{tabular}{lccc}
      \toprule 
        & \bfseries Cosine & \bfseries Michalewicz & \bfseries Modified Michalewicz\\ 
      \midrule 
      EI  & 0.49 & 0.52  & 0.68 \\
      TS & 0.55 & 0.53  & 0.63 \\
      V-UCB & 1.36  & 1.28 & 1.50 \\
      GP-UCB & 1.36 & 1.28 & 1.50 \\
      LW-UCB & 4.19 & 3.94 & 4.51 \\
      \bottomrule 
    \end{tabular}
\end{table*}

\begin{figure*}[!ht]
\centering
\subfloat[Cosine]{\includegraphics[width=0.32\textwidth]{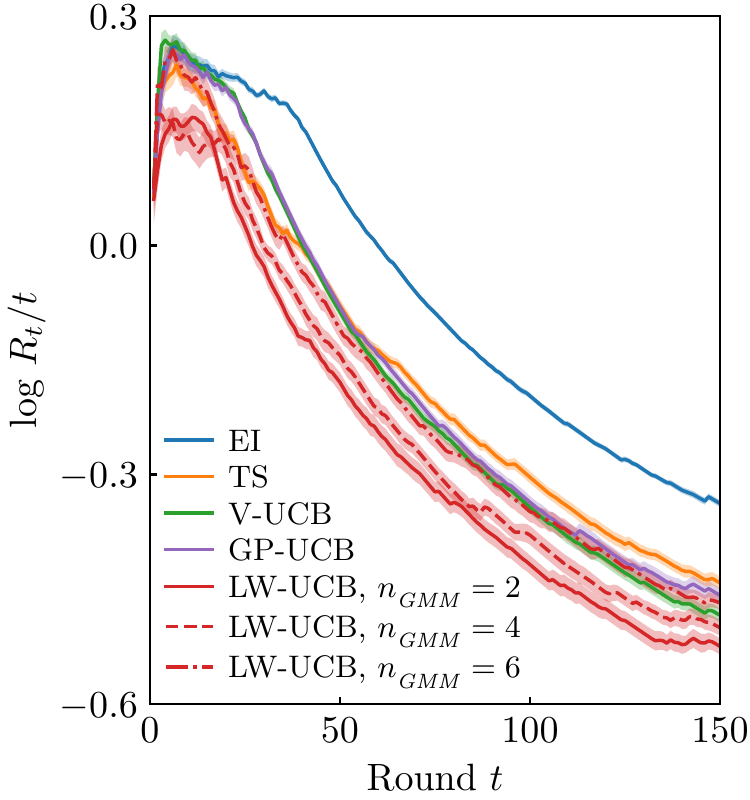}} ~~
\subfloat[Michalewicz]{\includegraphics[width=0.32\textwidth]{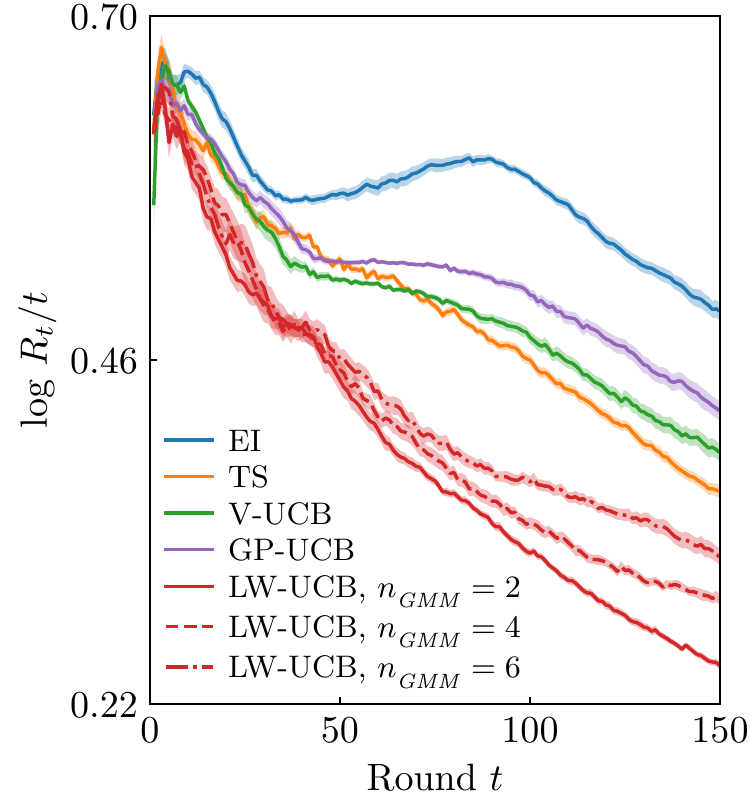}} ~~
\subfloat[Modified Michalewicz]{\includegraphics[width=0.32\textwidth]{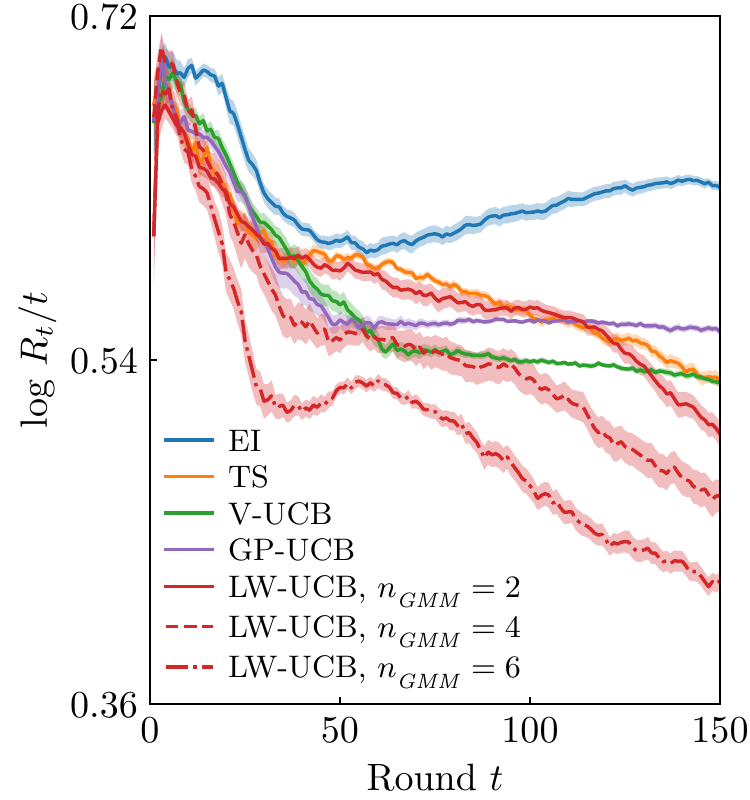}}

\caption{For the synthetic functions in Section \ref{sec:classic_functions}, performance of LW-UCB with various values of $n_\textit{GMM}$ compared to the other acquisition functions considered in this work.}
\label{sup:gmm}
\end{figure*}

We have also investigated the sensitivity of the LW-UCB criterion to the size of the Gaussian mixture model used in the approximation of the likelihood ratio. For the three synthetic functions \eqref{equ:Cosine_function}--\eqref{equ:Modified_Michalewicz_function}, we repeated the experiments with two additional values of $n_\textit{GMM}$.  Figure \ref{sup:gmm} shows that the performance of LW-UCB is essentially independent of the number of Gaussian components used in \eqref{equ:GMMApproximation} when the latent function is relatively simple, and that larger values of $n_\textit{GMM}$ are preferable when the complexity of the landscape grows and the number of optimal regions increases.

\subsection{A systematic study: wheel bandits}
\label{sec:wheel_bandits}

\begin{figure*}[t]
\centering
\subfloat[$\rho = 0.5$]{\includegraphics[width=\textwidth]{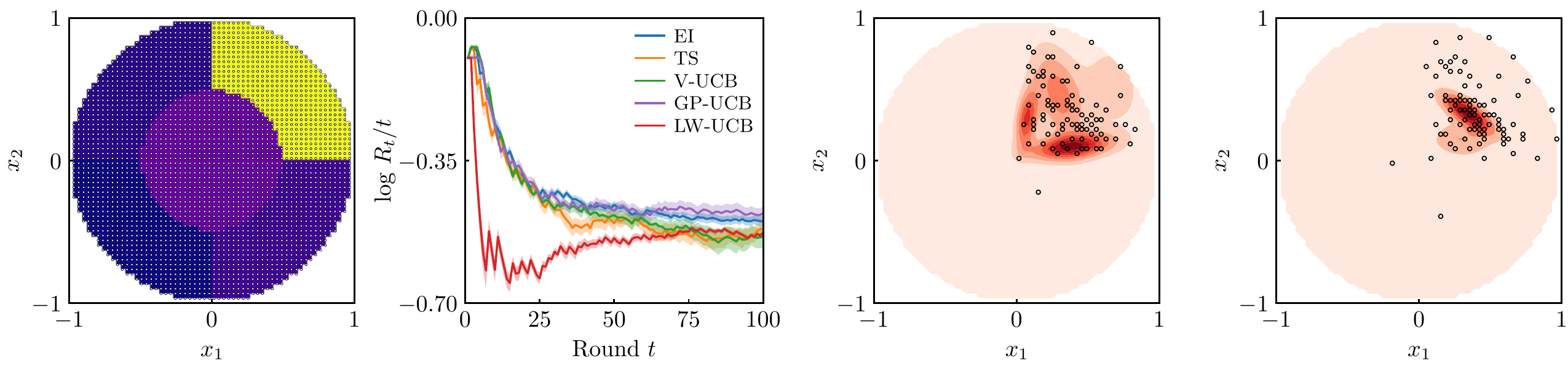}}

\subfloat[$\rho = 0.7$]{\includegraphics[width=\textwidth]{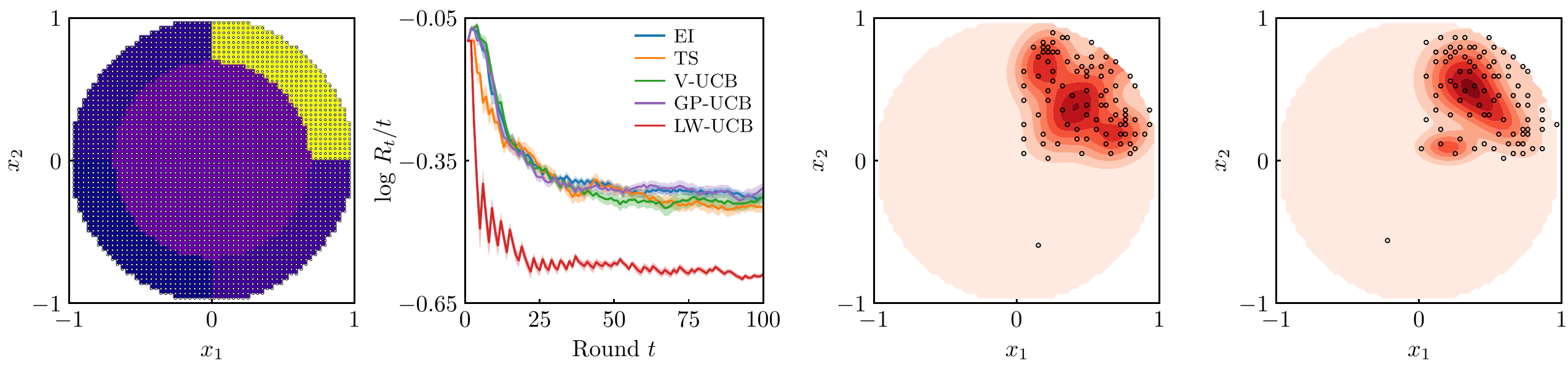}}

\subfloat[$\rho = 0.9$]{\includegraphics[width=\textwidth]{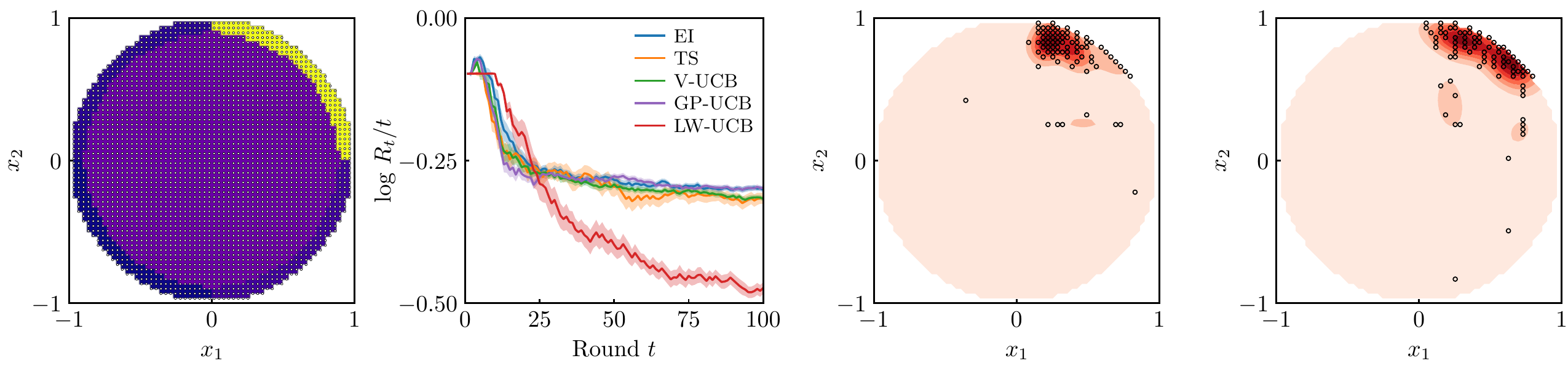}}

\caption{{\em Wheel bandit problem.} From left to right: locations of the bandits (white circles) and associated rewards (background color); cumulative regret for various acquisition functions; and for two representative trials of LW-UCB, distribution of the likelihood ratio (background color) learned by the GP model from the visited bandits (open circles) after $t=100$ rounds.}
\label{fig:Wheel}
\end{figure*}

In this section we consider a variant of the contextual wheel bandit problem discussed in \cite{riquelme2018deep}. The feasible domain is the unit disk ($0 \leq r \leq 1$) which is divided into five disjoint sectors.  The inner disk ($ 0 \leq r \leq \rho$) is sub-optimal with reward 0.2.  The upper left, lower right, and lower left quadrants of the outer ring ($\rho \leq r \leq 1$) are also sub-optimal, with rewards 0.05, 0.1, and 0, respectively (Figure \ref{fig:Wheel}).  The optimal bandits are located in the upper right quadrant of the outer ring and return a reward of 1, significantly higher than the other quadrants.  The parameter $\rho$ determines the difficulty of the problem.  For small $\rho$, the optimal region accounts for a large fraction of the domain, while for large $\rho$ the difficulty significantly increases.  We generate the bandits on a $70 \times 70$ uniform grid and retain those lying inside the unit disk.  Each bandit produces noisy rewards with $\sigma_n = 10^{-3}$. 

For $n_\textit{GMM} = 4$, Figure \ref{fig:Wheel} shows that the proposed LW-UCB criterion leads to significant gains in performance compared to conventional acquisition functions, especially as the value of $\rho$ increases and the optimal bandits become scarcer.  Figure \ref{fig:Wheel} also shows that the attention mechanism embedded in the likelihood ratio encourages exploration of the extreme-reward region.  It is also interesting to note that in all cases investigated, the expected improvement, Thompson sampling, V-UCB, and GP-UCB deliver nearly identical performance, even in the asymptotic regime, unlike LW-UCB which provides consistently faster convergence.

\subsection{Spatio-temporal environment monitoring with sensor networks}
\label{sec:sensors_example}
\textcolor{black}{Finally, we demonstrate the approach using two real-world datasets: the temperature dataset considered in  \cite{srinivas2009gaussian} and the air quality dataset considered in \cite{cai2021periodic}.}

\subsubsection{Temperature dataset}
\label{sec:temperature_data}

\begin{figure*}[!ht]
\centering
\subfloat[][\label{fig:tempa}]{\includegraphics[height=0.25\textwidth]{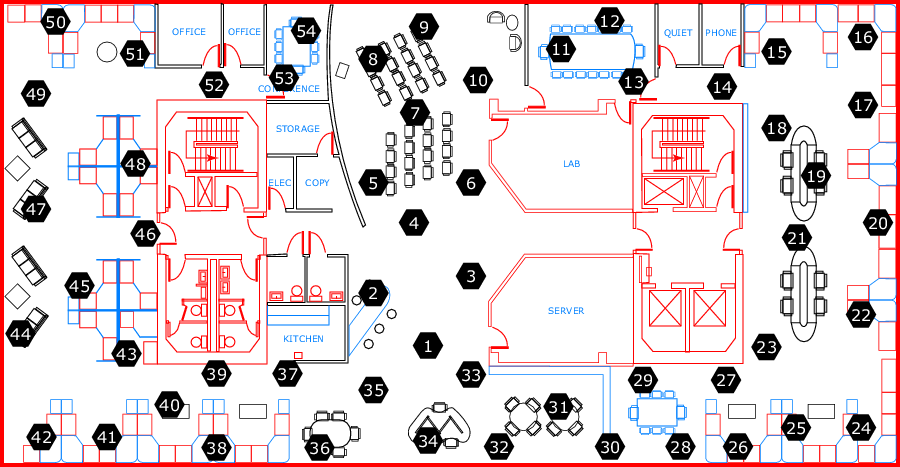}} \qquad\qquad
\subfloat[][\label{fig:tempb}]{\includegraphics[height=0.25\textwidth]{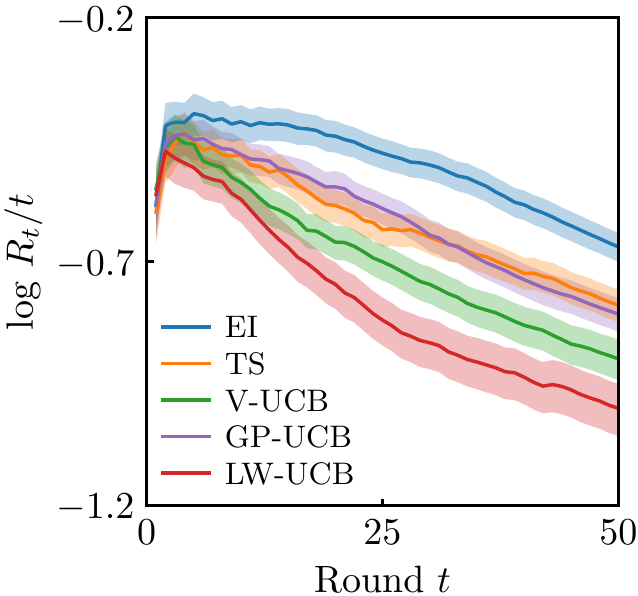}}

\subfloat[][\label{fig:tempc}]{\includegraphics[width=0.49\textwidth]{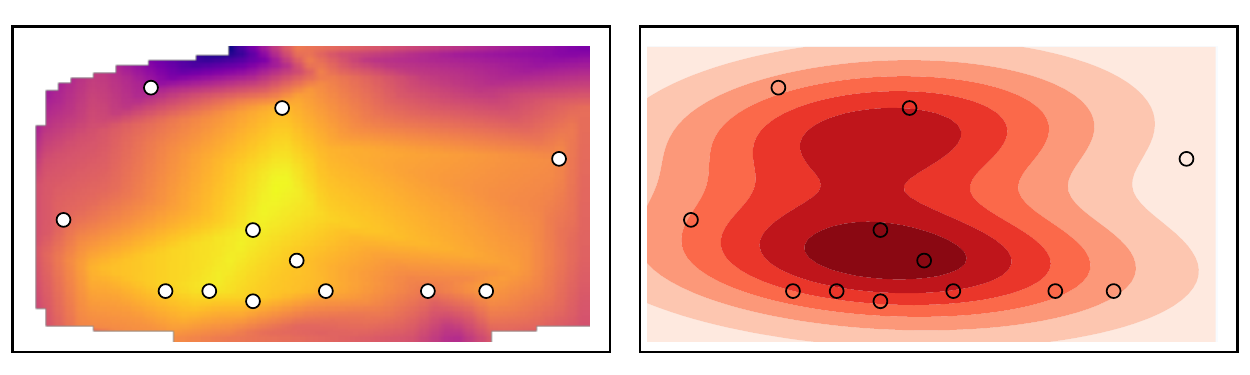}} ~~
\subfloat[][\label{fig:tempd}]{\includegraphics[width=0.49\textwidth]{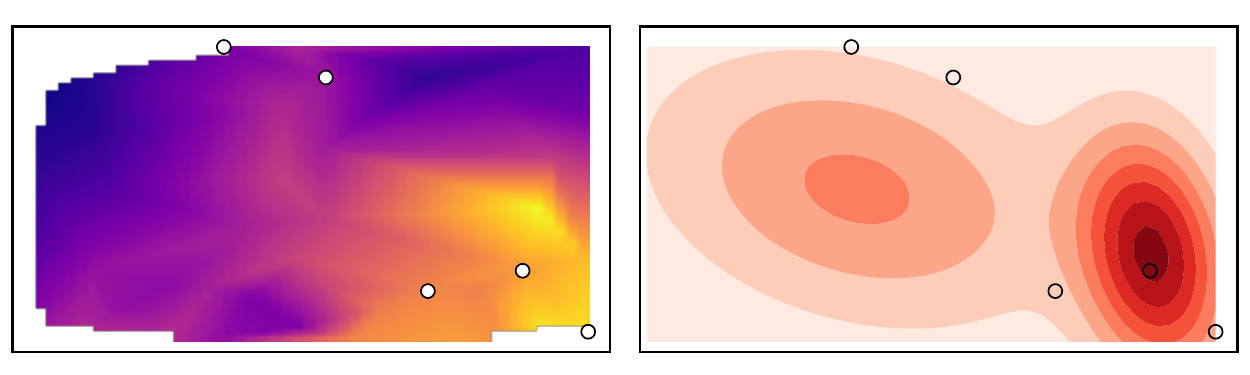}}

\subfloat[][\label{fig:tempe}]{\includegraphics[width=0.49\textwidth]{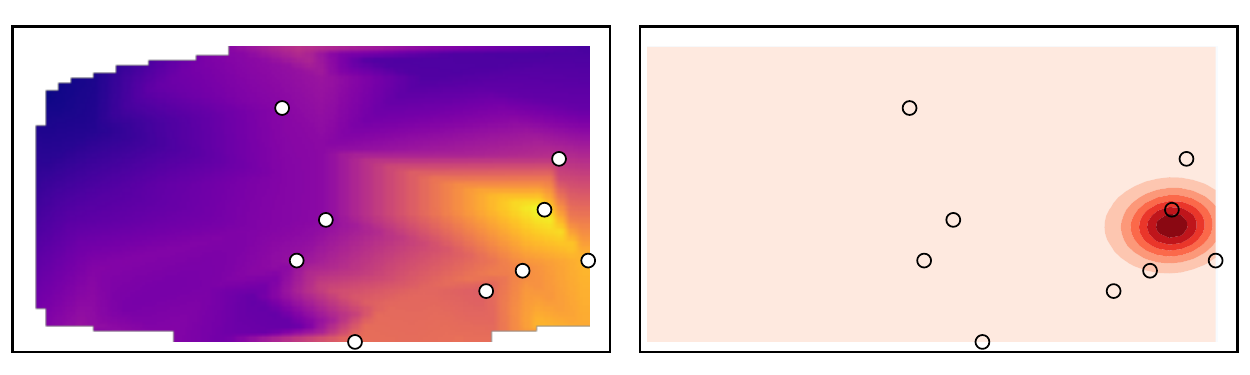}} ~~
\subfloat[][\label{fig:tempf}]{\includegraphics[width=0.49\textwidth]{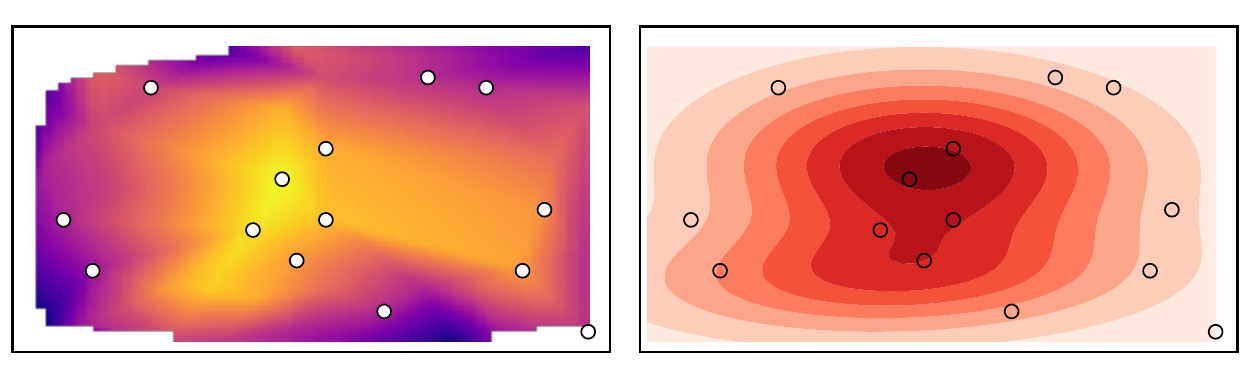}}

\subfloat[][\label{fig:tempg}]{\includegraphics[width=0.49\textwidth]{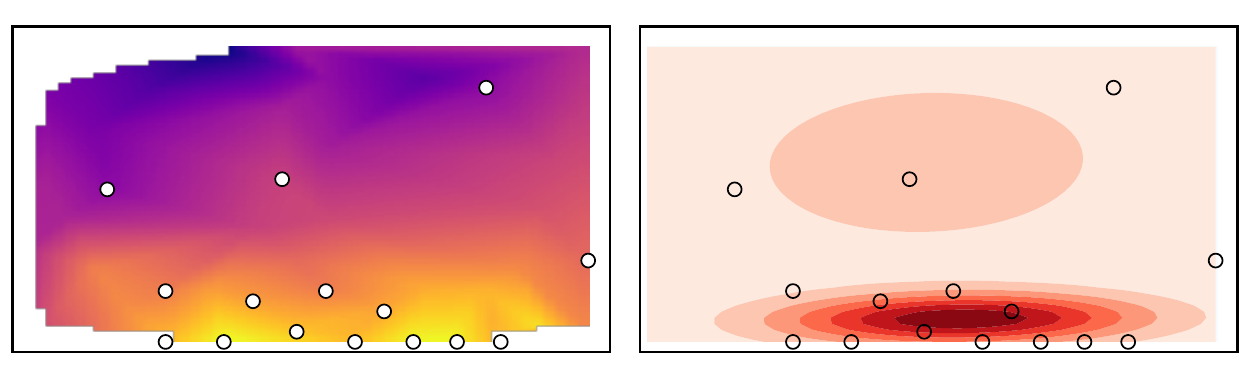}} ~~
\subfloat[][\label{fig:temph}]{\includegraphics[width=0.49\textwidth]{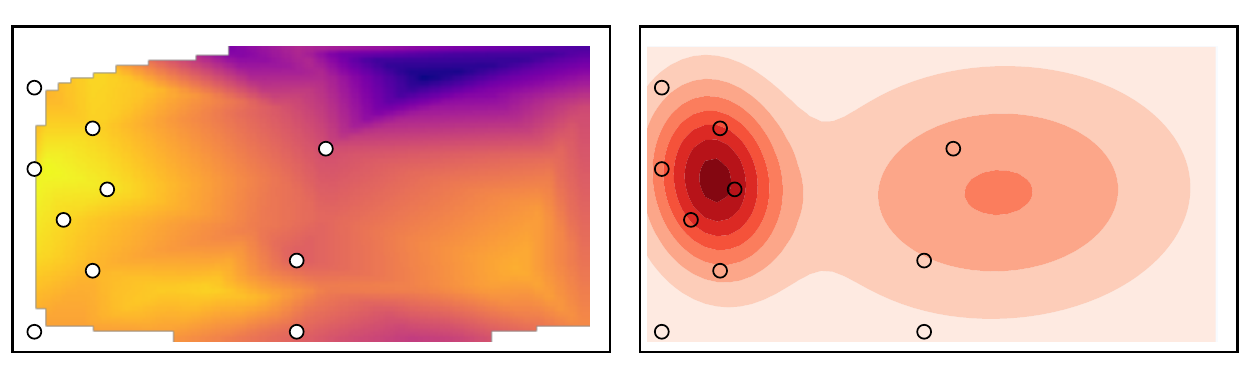}}

\caption{{\em Spatio-temporal monitoring of room temperature with sensor networks.} (a) Sensor locations; (b)  cumulative regret for various acquisition functions; and (c--h) for six representative trials of LW-UCB, spatial distribution of temperature (left panel) and the likelihood ratio (right panel) learned by the GP model from the activated sensors (circles) after $t=50$ rounds.}
\label{fig:Temperature_data}
\end{figure*}

The temperature dataset\footnote{\url{http://db.csail.mit.edu/labdata/labdata.html}} contains temperature measurements collected by 46 sensors deployed in the Intel Berkeley Research lab (Figure \ref{fig:tempa}).  As in \cite{srinivas2009gaussian}, our goal is to find locations of highest temperature by sequentially activating the available sensors while using as few sensor switches as possible in order to save electric power.  Our working dataset consists of 500 temperature snapshots collected every ten minutes over a three-day period.  For each temperature snapshot, we initialize the algorithm by randomly activating $n=3$ sensors.  The sensors (i.e., the bandits) produce rewards that are corrupted by small Gaussian noise with $\sigma_n=10^{-4}$.  We use $n_\textit{GMM} = 2$ for the GMM approximation of the likelihood ratio.

For this real-world problem, Figure \ref{fig:tempb} shows that LW-UCB performs better than the other acquisition schemes. Figures \ref{fig:tempc}--\ref{fig:temph} show that the likelihood ratio draws the algorithm's attention to the bandits whose rewards are high by artificially inflating the model uncertainty for these bandits. We note that, in contrast to the examples considered previously, here the bandits are few and far between.  For instance, there is no sensor data available in the server room and the stairwell (see Figure \ref{fig:tempa}). Because of the sparsity of the data, finding the best sensor to activate is more challenging.  But this does not seem to negatively affect the LW-UCB acquisition criterion, which is able to identify and explore the relevant areas more intelligently than the other acquisition functions.

\subsubsection{Air quality dataset}
\label{sec:pollution_data}
\textcolor{black}{The air quality dataset\footnote{\url{https://datos.madrid.es/portal/site/egob}} contains concentration measurements of pollutants and other particles collected by 24 sensors deployed in the Madrid metropolitan area. Each sensor is uniquely identified by its longitude, latitude, and elevation.  We focus on finding the locations of highest nitrogen dioxide (NO$_2$) by sequentially activating the available sensors to identify the region of worst air quality. The parsed data consists of 200 pollution snapshots collected every hour over a 10-day period in March 2018. For each pollution snapshot, we initialize the algorithm by randomly activating $n=3$ sensors.  As in the temperature example, the rewards produced by each sensor are corrupted by small Gaussian noise with $\sigma_n=10^{-4}$. We use $n_\textit{GMM} = 2$ for the GMM approximation of the likelihood ratio.}

\begin{figure}[!ht]
\centering
\includegraphics[height=0.75\textwidth]{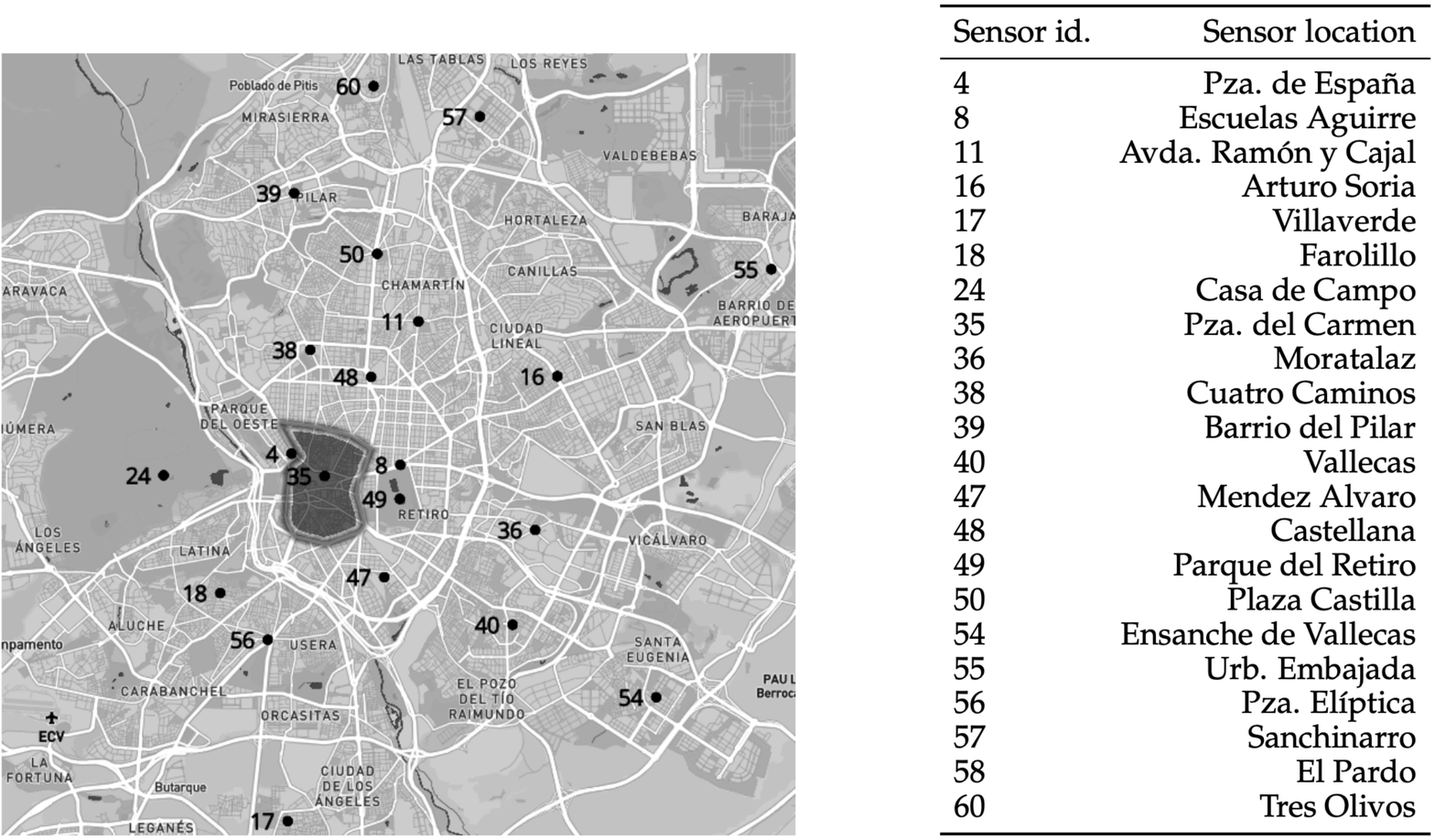}
\caption{{\em Spatio-temporal monitoring of air quality with sensor networks:} Locations of 22 out of the 24 sensors \textcolor{black}{(i.e. excluding sensors \#27 (Barajas Pueblo) and \#59 (Juan Carlos I))} used to monitor air quality in the Madrid metropolitan area (original figure from \cite{lebrusan2020using}).}
\label{fig:Pollution_data_sensor_locations}
\end{figure}

\textcolor{black}{
We consider two cases: the partial-context case in which we only use longitude and latitude as the contextual information for each sensor (we summarize the locations for the sensors in Figure \ref{fig:Pollution_data_sensor_locations}); and the full-context case in which elevation is also accounted for.  The partial-context case is worthy of investigation because, for the geographical area considered, the effect of elevation on NO$_2$ concentration should be quite small. Also, the number of sensors is relatively small (24), so using partial contextual information allows us to reduce the rank of the problem.}

\begin{figure*}[!ht]
\centering

\subfloat[][\label{fig:pola}]{\includegraphics[height=0.25\textwidth]{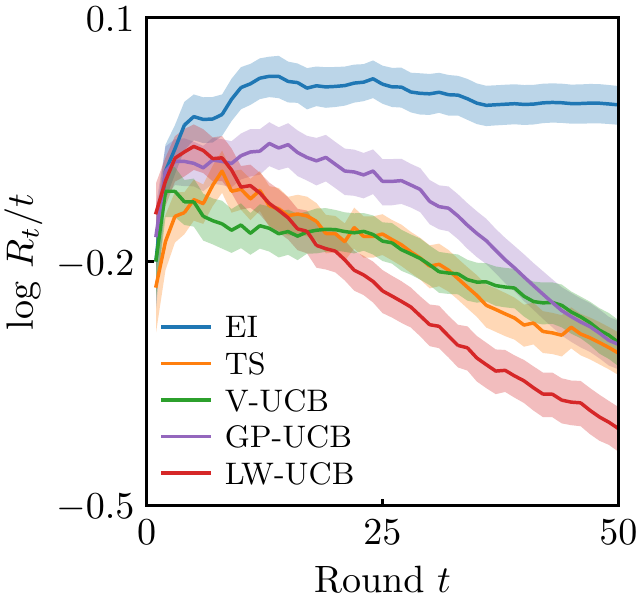}}  \qquad
\subfloat[][\label{fig:polb}]{\includegraphics[height=0.25\textwidth]{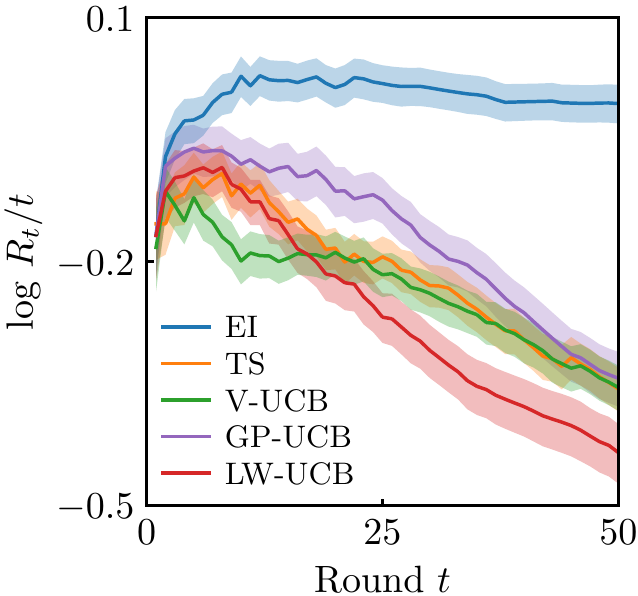}} 

\subfloat[][\label{fig:polc}]{\includegraphics[width=0.32\textwidth]{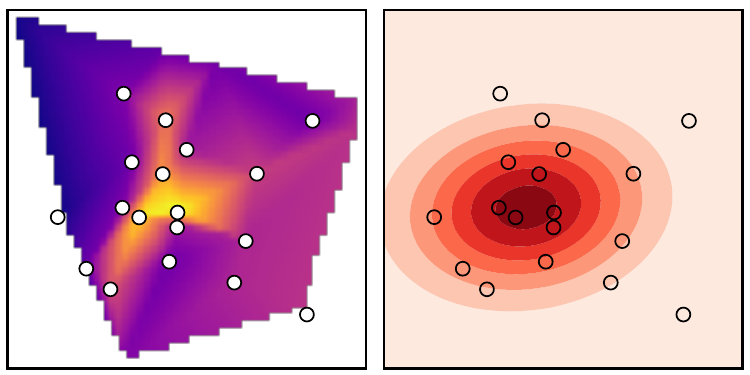}} ~~
\subfloat[][\label{fig:pold}]{\includegraphics[width=0.32\textwidth]{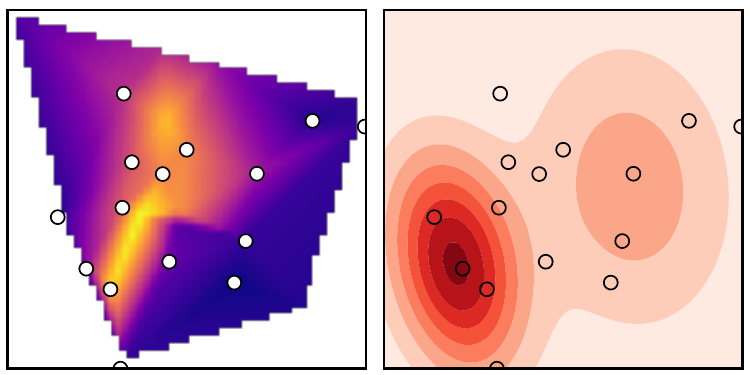}} ~~
\subfloat[][\label{fig:pole}]{\includegraphics[width=0.32\textwidth]{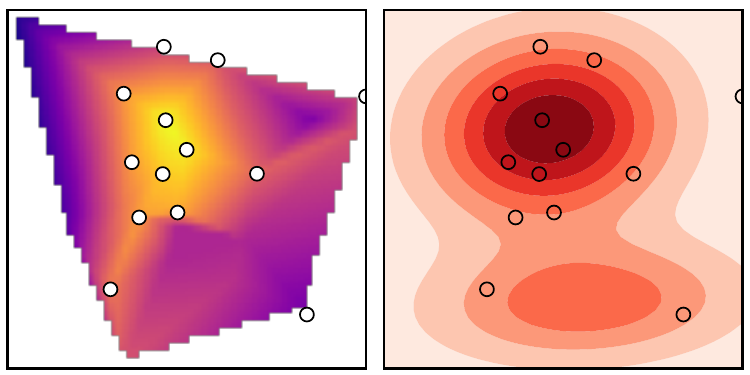}} 

\subfloat[][\label{fig:polf}]{\includegraphics[width=0.32\textwidth]{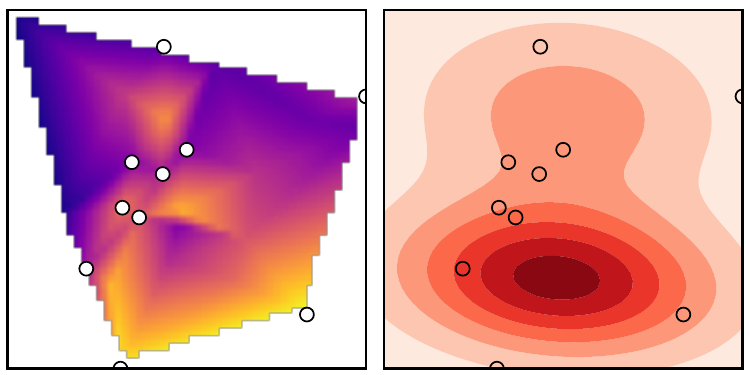}} ~~
\subfloat[][\label{fig:polg}]{\includegraphics[width=0.32\textwidth]{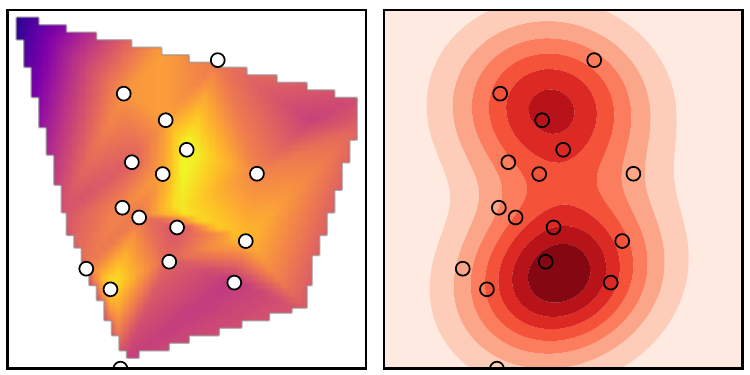}} ~~
\subfloat[][\label{fig:polh}]{\includegraphics[width=0.32\textwidth]{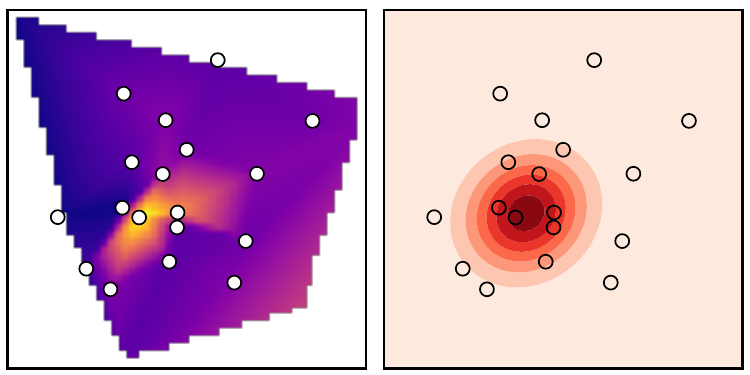}} 

\caption{{\em Spatio-temporal monitoring of air quality with sensor networks.} Cumulative regret for (a) the partial-context case and (b) the full-context case; and (c--h) for the partial-context case and six representative trials of LW-UCB, spatial distribution of NO$_2$ concentration (left panel) and the likelihood ratio (right panel) learned by the GP model from the activated sensors (circles) after $t=50$ rounds.}
\label{fig:Pollution_data}
\end{figure*}

\textcolor{black}{For both cases, Figures \ref{fig:pola} and \ref{fig:polb} show that LW-UCB outperforms the other acquisition functions considered. The snapshots shown in figures \ref{fig:polc}--\ref{fig:polh} reinforce the utility of the likelihood ratio to identify regions of high NO$_2$ concentration (i.e., poor air quality) more efficiently.  As in the temperature example, the sparsity of the data does not seem to hamper the ability of LW-UCB to converge to the optimal bandits faster than the other acquisition schemes.}

\section{Conclusions}\label{sec:discussion}

We have proposed a novel output-weighted acquisition function (LW-UCB) for sequential decision making. Our approach leverages the information provided by the GP regression model to regularize uncertainty and favor exploration of abnormally large payoff values.  The regularizer takes the form of a sampling weight---the likelihood ratio---and can be efficiently approximated by a Gaussian mixture model. The likelihood ratio provides a principled way to balance exploration and exploitation in multi-armed bandit optimization problems where the goal is to maximize the cumulative reward. The benefits of the proposed method have been systematically established via several benchmark examples which demonstrated superiority of our method compared to classical acquisition functions (expected improvement, Thompson sampling, and two variants of UCB).  

Though the proposed LW-UCB criterion yields superior performance in bandit problems, several questions remain open.  First, a theoretical analysis of the convergence behavior of LW-UCB is needed, in the same way that information gain has helped characterize the convergence of GP-UCB \cite{srinivas2009gaussian, krause2011contextual}.  The second avenue is to investigate more complex cases with high-dimensional contexts and multi-output GP priors. The latter can be readily accommodated in our JAX implementation which leverages automatic differentiation to allow efficient gradient-based optimization of the LW-UCB criterion for arbitrary GP priors.  The third question has to do with extending the proposed approach to other Bayesian inference schemes, e.g., Bayesian linear regression \cite{chu2011contextual}, Bayesian neural networks \cite{riquelme2018deep}, and variational inference \cite{hoffman2013stochastic}.  Finally, there is the question of how to adapt the proposed framework for use in more general Markov decision processes and reinforcement learning problems \cite{sutton2018reinforcement} where contextual information is typically high-dimensional and rewards are obtained after multiple trials rather than instantaneously.

\section*{Acknowledgements}
Y.Y. and P.P. acknowledge support from the DOE grant DE-SC0019116, AFOSR grant FA9550-20-1-0060, and DOE-ARPA grant DE-AR0001201. A.B. and T.S. would like to thank the support from the AFOSR-MURI Grant No. FA9550- 21-1-0058 and the ONR Grant No. N00014-21-1-2357.

\bibliographystyle{unsrt}
\bibliography{main.bib}

\end{document}